\documentclass[preprint,12pt]{elsarticle}

\usepackage{amssymb}
\usepackage[12pt]{extsizes}
\usepackage[utf8]{inputenc}
\usepackage{amsmath}
\usepackage{amssymb}
\usepackage{amsthm}
\usepackage{tabularx}
\usepackage{graphics}
\usepackage{graphicx}
\usepackage[font=small,labelfont=bf,margin=0.39in]{caption}
\usepackage[dvipsnames]{xcolor}
\usepackage[hidelinks]{hyperref}
\hypersetup{
    colorlinks = true,
    linkcolor={blue!50!black},
    citecolor={blue!50!black},
    urlcolor={blue!80!black}
}
\usepackage{geometry}
\usepackage{mathtools}
\usepackage{amsfonts}
\usepackage{relsize}
\usepackage{multirow}
\usepackage{booktabs}
\usepackage{tabularx}
\usepackage{listings}
\usepackage{enumitem}
\usepackage{setspace}

\definecolor{purple}{RGB}{250,000,180}

\def\Hbb{\mathbb{H}}

\usepackage{xspace}
\usepackage[ruled,vlined]{algorithm2e}

\usepackage{multirow}

\newcommand{\ie}{\emph{i.e.}\xspace}
\newcommand{\etal}{\emph{et al.}\xspace}

\newtheorem*{problem*}{Problem}

\newcommand{\xhdr}[1]{\vspace{1mm}\noindent{{\bf #1.}}}
\newcommand{\hide}[1]{}

\usepackage[scale=.95,type1]{cabin}
\usepackage[zerostyle=c,scaled=.94]{newtxtt}
\usepackage[cal=boondoxo]{mathalfa}
\usepackage{microtype}

\usepackage{etoolbox}
\apptocmd{\thebibliography}{\raggedright}{}{}

\usepackage{amsmath,amsfonts,bm}

\def\eqref#1{equation~\ref{#1}}

\DeclareMathAlphabet{\mathsfit}{\encodingdefault}{\sfdefault}{m}{sl}
\SetMathAlphabet{\mathsfit}{bold}{\encodingdefault}{\sfdefault}{bx}{n}

\def\bH{\mathbf{H}}

\begin{document}

\begin{frontmatter}



\title{Multimodal Learning on Graphs for Disease Relation Extraction
}
\author[1,2]{Yucong Lin\fnref{coauther}}
\author[3]{Keming Lu\fnref{coauther}}
\author[4,5]{Sheng Yu}
\author[6,7]{Tianxi Cai}
\author[7,8,9]{Marinka Zitnik\corref{cor}}
\fntext[coauther]{Equal contribution}
\cortext[cor]{Corresponding author: marinka@hms.harvard.com; 1-617-432-5138; Department of Biomedical Informatics, Harvard Medical School, 10 Shattuck Street, Boston, MA 02115, USA}


\affiliation[1]{organization={Institute of Engineering Medicine, Beijing Institute of Technology},
            addressline={Beijing, China}, 
            city={Beijing},
            country={China}}

\affiliation[2]{organization={Beijing Engineering Research Center of Mixed Reality and Advanced Display, School of Optics and Photonics,Beijing Institute of Technology},
            city={Beijing},
            country={China}}

\affiliation[3]{organization={Viterbi School of Engineering, University of Southern California},
            city={Los Angeles},
            state={CA},
            postcode={90007},
            country={USA}}

\affiliation[4]{organization={Center for Statistical Science, Tsinghua University},
            city={Beijing},
            country={China}}
            
\affiliation[5]{organization={Department of Industrial Engineering, Tsinghua University},
            city={Beijing},
            country={China}}

\affiliation[6]{organization={Department of Biostatistics, Harvard T.H.Chan School of Public Health},
            city={Boston},
            state={MA},
            postcode={02115},
            country={USA}}
            
\affiliation[7]{organization={Department of Biomedical Informatics, Harvard Medical School},
            city={Boston},
            state={MA},
            postcode={02115},
            country={USA}}

\affiliation[8]{organization={Broad Institute of MIT and Harvard},
            city={Boston},
            state={MA},
            postcode={02142},
            country={USA}}

\affiliation[9]{organization={Harvard Data Science Initiative},
            city={Cambridge},
            state={MA},
            postcode={02138},
            country={USA}}

\begin{abstract}
\noindent \textbf{Objective:} Disease knowledge graphs are a way to connect, organize, and access disparate information about diseases with numerous benefits for artificial intelligence (AI). To create knowledge graphs, it is necessary to extract knowledge from multimodal datasets in the form of relationships between disease concepts and normalize both concepts and relationship types.
\\[1mm]
\noindent \textbf{Methods:} We introduce REMAP, a multimodal approach for disease relation extraction and classification. The REMAP machine learning approach jointly embeds a partial, incomplete knowledge graph and a medical language dataset into a compact latent vector space, followed by aligning the multimodal embeddings for optimal disease relation extraction. 
\\[1mm]
\noindent \textbf{Results:} We apply REMAP approach to a disease knowledge graph with 96,913 relations and a text dataset of 1.24 million sentences. 
On a dataset annotated by human experts, REMAP improves text-based disease relation extraction by 10.0\% (accuracy) and 17.2\% (F1-score) by fusing disease knowledge graphs with text information. Further, REMAP leverages text information to recommend new relationships in the knowledge graph, outperforming graph-based methods by 8.4\% (accuracy) and 10.4\% (F1-score). \\[1mm]
\noindent \textbf{Conclusion:} REMAP is a multimodal approach for extracting and classifying disease relationships by fusing structured knowledge and text. REMAP provides a flexible neural architecture to easily find, access, and validate AI-driven relationships between disease concepts. \\[1mm]
\end{abstract}

\begin{graphicalabstract}
\includegraphics[width=\linewidth]{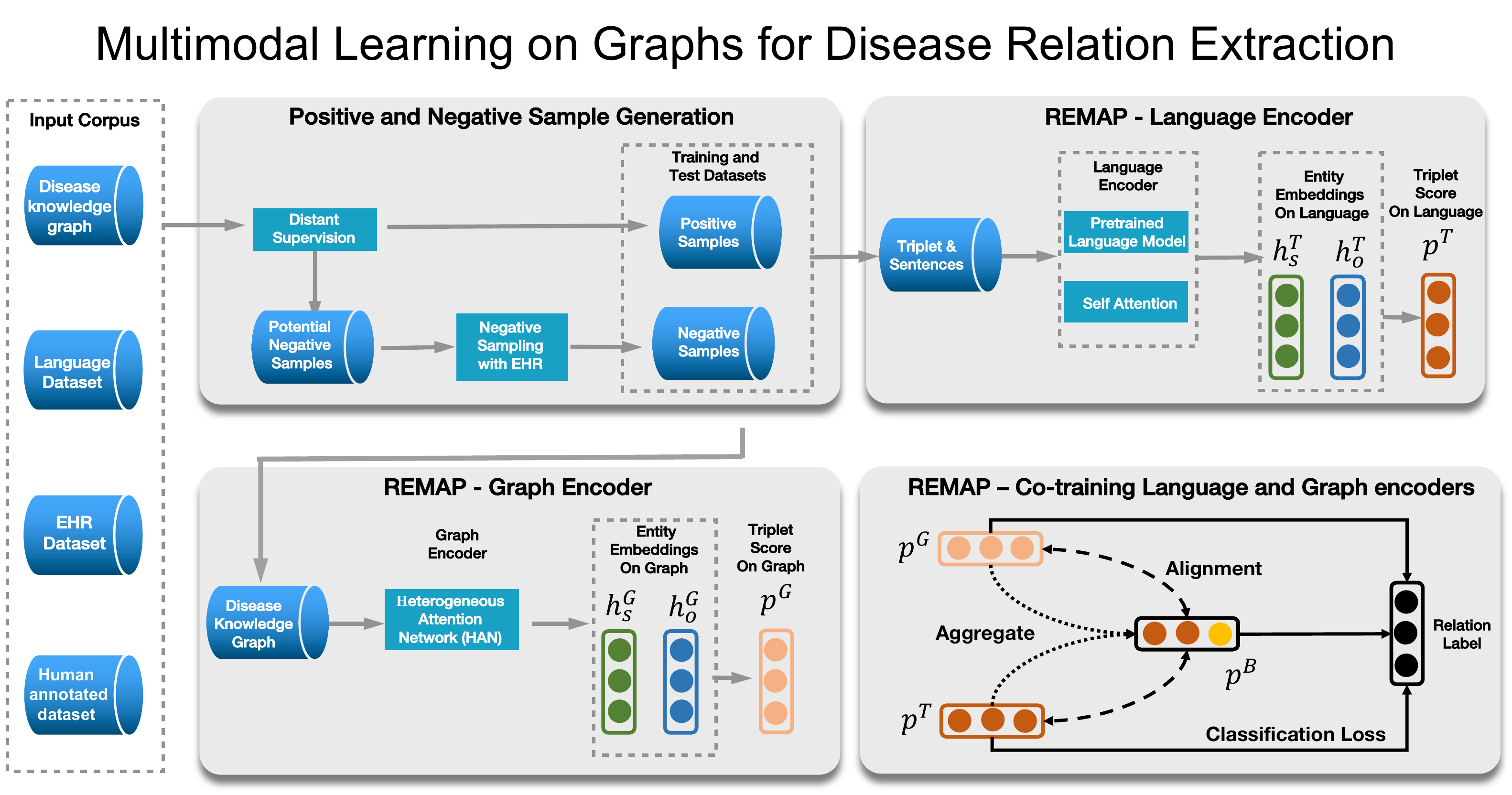}
\end{graphicalabstract}

\begin{highlights}
\item We develop a flexible multimodal approach for extracting and classifying diverse kinds of disease-disease
relationships (REMAP). REMAP fuses knowledge graph embeddings with deep language models and can flexibly accommodate missing data types.
\item Evaluation against a clinical expert annotated dataset shows that REMAP achieves 88.6\% micro-accuracy and 81.8\% micro-F1 score, outperforming text-based methods by 10 and 17.2 percentage points, respectively.
\item  We release a high-quality test
dataset of gold-standard annotations developed as a consensus of three clinical experts for evaluating disease-disease relation extraction, together with the open-source implementation of REMAP. 
\end{highlights}


\end{frontmatter}


\section{Introduction}\label{sec:intro}
Disease knowledge graphs are a way to connect, organize, and access disparate data and information resources about diseases with numerous benefits for artificial intelligence (AI).
Systematized knowledge can be injected into AI methods to imitate human experts' reasoning so that AI-driven hypotheses can be easily found, accessed, and validated. For example, disease knowledge graphs (KGs) power AI applications, such as identification of disease treatments~\cite{ruiz2021identification} and electronic health record (EHR) retrieval~\cite{hong2021clinical}. However, creating high-quality knowledge graphs requires extracting relationships between diseases from disparate information sources, such as free text in the EHRs and semantic knowledge representation from literature.

Traditionally, KGs were constructed via manual efforts, requiring humans to input every fact~\cite{bollacker2008freebase}. In contrast, rule-based~\cite{rindflesch2011semantic} and semi-automated~\cite{carlson2010toward,dong2014knowledge} methods, while scalable, can suffer from poor accuracy and low recall rates. As a result, an enticing alternative is to create KGs by extracting relationships from literature and building large-scale KGs that comprehensively cover a domain of interest. These methods leverage pre-trained language models~\cite{devlin2018bert,scibert} and have advanced the analysis of biomedical knowledge graphs~\cite{MTB,zhong2020frustratingly,lin2020highthroughput,chen2022biomedical,li2017noise}. Another approach for populating KGs with relations uses knowledge graph embeddings (KGE), which directly predict new relations in partial, incomplete knowledge graphs. KGE methods learn how to represent every entity (i.e., node) and relation (i.e., edge) in a graph as a distinct point in a low-dimensional vector space (i.e., embedding) so that performing algebraic operations in this learned space reflects the topology of the graph~\cite{wang2017knowledge}. Embeddings produced by KGE methods can be remarkably powerful for downstream AI applications~\cite{li2021representation,zitnik2016collective,shi2017proje,lin2015learning,wang2019kgat,sun2018recurrent}. Widely used KGE methods include translation models~\cite{bordes2013translating,wang2014knowledge,lin2015learning,ji2015knowledge}, bilinear models~\cite{nickel2011three,yang2014embedding,trouillon2017knowledge,balavzevic2019tucker}, and graph neural networks (GNNs)~\cite{schlichtkrull2018modeling,busbridge2019relational,wang2019heterogeneous,zitnik2018modeling}. These methods leverage embeddings to predict new relations, thereby completing sparse knowledge areas and systematically growing an existing KG. However, extracting relations from a single data type may suffer from bias, noise, and incompleteness. For example, in language-based methods, the training dataset is collected using distant supervision~\cite{mintz2009distant}, which creates noisy sentences that can mislead relation extraction. Further, graph-based methods can suffer from out-of-dictionary problems, which limit the ability to model relations involving entities previously not in the KG~\cite{bordes2013translating,balazevic2019tucker}.

Nevertheless, language-based and graph-based methods both have advantages. For example, language-based methods can reason over large datasets created using techniques such as distant supervision and contrastive learning, and graph-based methods can operate on noisy and incomplete knowledge graphs, providing robust predictions. An emerging strategy to advance relation extraction thus leverages multiple data types simultaneously~\cite{liu2020k,sun2019ernie,zhang2019ernie,he2019integrating,koncel2019text,sun2020colake,hu2019improving,xu2019connecting,zhang2019long,wang2020model,wang2014knowledge,han2016joint,ji2020joint,dai2019distantly,stoica2020improving} with multimodal learning~\cite{zitnik2019machine}, outperforming rule-based~\cite{rindflesch2011semantic} and semi-automated~\cite{carlson2010toward,dong2014knowledge} methods. However, existing approaches are limited in two ways, which we outline next.

First, KGs provide only positive samples (i.e., examples of true disease-disease relationships), while existing methods also require negative samples that, ideally, are disease pairs that resemble positive samples but are not valid relationships. Methods for positive-unlabeled~\cite{he2020improving} and contrastive~\cite{le2020contrastive,su2021improving} learning can address this challenge by sampling random disease pairs from the dataset as negative proxy samples and ensuring a low false-positive rate. However, these methods may not generalize well in real-world applications because random negative samples are not necessarily realistic and fail to represent the boundary cases. To improve the quality of negative sampling in disease relation extraction, we introduce an EHR-based negative sampling strategy in this work. With the strategy, our approach generates negative samples using disease pairs that rarely appear together in EHRs, thus having realistic negative samples to enable the broad generalization of the approach.

Second, only graph or language information is available for a subset of diseases but not both modalities.
For example, in Lin \etal~\cite{lin2020highthroughput}, over 60\% disease pairs in the KG had no corresponding text information; there were also cases with text but no graph information. Thus, multimodal approaches must be flexible, meaning they can make predictions when only one data type is available. Unfortunately, a small number of existing multimodal approaches with such capability~\cite{wang2020multimodal,suo2019metric,zhou2019latent,yang2018semi} do not consider language and graphs. 
Further, some studies~\cite{cai2018deep,jaques2017multimodal} use adversarial learning to impute data from missing modalities, but the imputed values can introduce unwanted bias, leading to distribution shifts. To address this issue, in this work, we develop a multimodal de-coupled architecture where language and graph modules interact only through shared parameters and a cross-modal loss function. This approach ensures our model can take advantage of both language and graph inputs and identify disease relations using either single or multimodal inputs.

\begin{figure}
\centering
\includegraphics[width=0.88\linewidth]{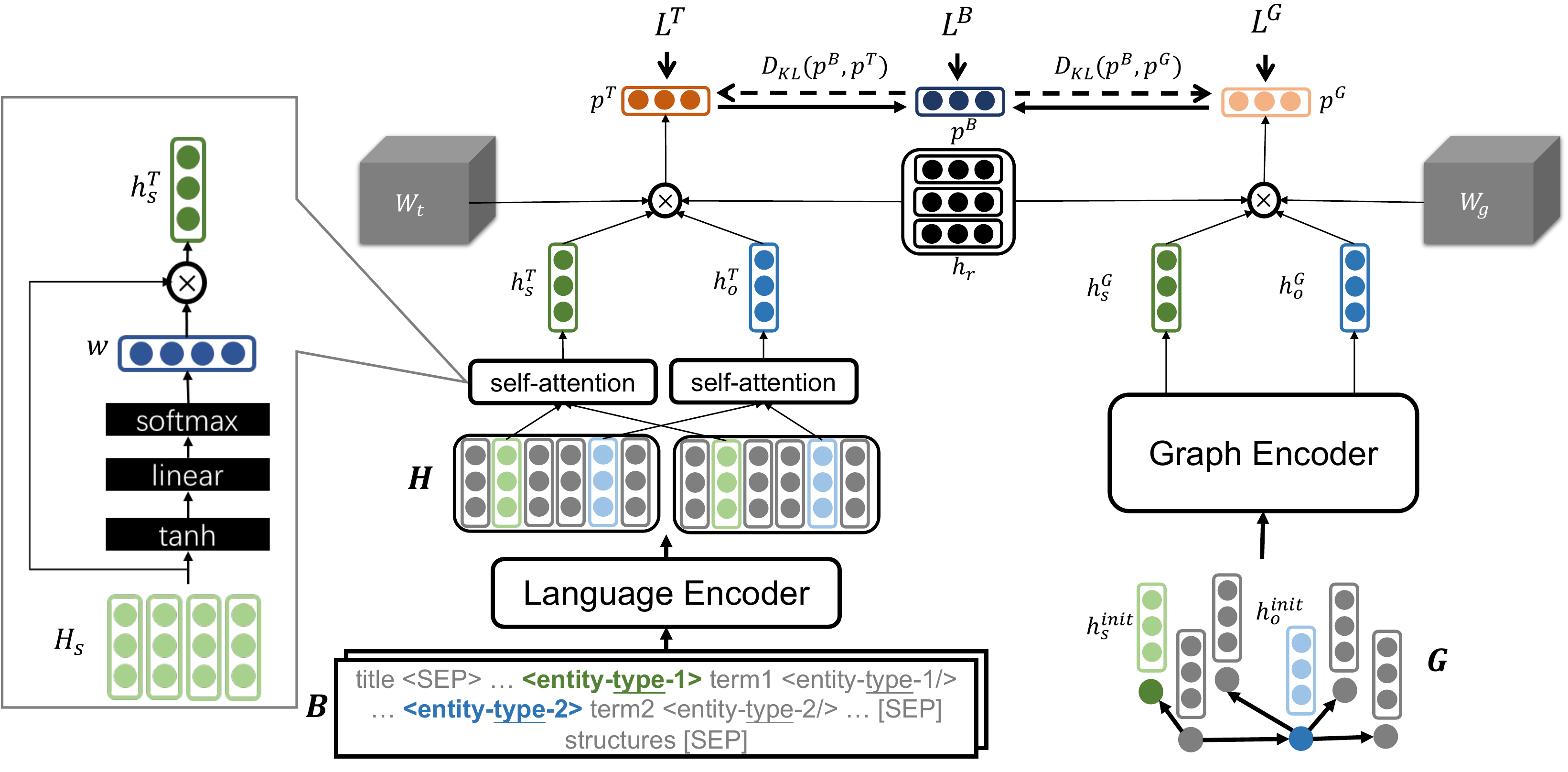} 
\caption{\textbf{Overview of REMAP architecture.} REMAP introduces a novel co-training learning strategy that continually updates a multimodal language-graph model for disease relation extraction and classification. Language and graph encoders specify deep transformation functions that embed disease concepts (i.e., subject entities $s$ and object entities $o$) from the language data $D^T$ and disease knowledge graph $D^G$ into compact embeddings, producing condensed summaries of language semantics and biomedical knowledge for every disease. Embeddings output by the encoders (i.e., $\mathbf{h}_s^T$, $\mathbf{h}_o^T$, $\mathbf{h}_o^G$, $\mathbf{h}_s^G$) are then combined in a disease relation type-specific manner (e.g., ``differential diagnosis'' and ''may cause'' relation types) and passed to a scoring function that calculates the probability representing how likely two diseases are related to each other and what kind of relationship exists between them.}
\label{Fig. 1}
\end{figure}

\xhdr{Present work}
We introduce REMAP (Relation Extraction with Multimodal Alignment Penalty)\footnote{Python implementation of REMAP is available on Github at \url{https://github.com/Lukeming-tsinghua/REMOD}. Our dataset of domain-expert annotations is at \url{https://doi.org/10.6084/m9.figshare.17776865}.}, a multimodal approach for extracting and classifying disease-disease relations (Figure~\ref{Fig. 1}). REMAP is a flexible multimodal algorithm that jointly learns over text and graphs with a unique capability to make predictions even when a disease concept exists in only one data type. To this end, REMAP specifies graph-based and text-based deep transformation functions that embed each data type separately and optimize unimodal embedding spaces such that they capture the topology of a disease KG or the text semantics of disease concepts. Finally, to achieve data fusion, REMAP aligns unimodal embedding spaces through a novel alignment penalty loss using shared disease concepts as anchors. This way, REMAP can effectively model data type-specific distribution and diverse representations while also aligning embeddings of distinct data types. Further, REMAP can be jointly trained on both graph and text data types but evaluated and implemented on either of the two modalities alone. In summary, the main contributions of this study are:
\begin{itemize}[leftmargin=*]
 \item We develop REMAP, a flexible multimodal approach for extracting and classifying disease-disease relations. REMAP fuses knowledge graph embeddings with deep language models and can flexibly accommodate missing data types, which is necessary to facilitate REMAP's validation and transition into biomedical implementation.
 \item We rigorously evaluate REMAP for extraction and classification of disease-disease relations. To this end, we create a training dataset using distant supervision and a high-quality test dataset of gold-standard annotations provided by three clinical domain experts. Evaluations show that REMAP achieves 88.6\%  micro-accuracy and 81.8\% micro-F1 score on the human-annotated dataset, outperforming text-based methods by 10 and 17.2  percentage points, respectively. Further, REMAP achieves the best performance, 89.8\% micro-accuracy, and 84.1\% micro-F1 score, surpassing graph-based methods by 8.4 and 10.4 percentage points, respectively.
\end{itemize}

\xhdr{JBI significance statement}
The significance of this study can be summarized as follows.

\begin{itemize}[leftmargin=*]
\item \textbf{Problem or issue:} Enhance disease relation extraction through multimodal learning from language information and knowledge graphs.

\item \textbf{What is already known:} Precise interpretation of disease-disease relationships is essential for building high-quality medical knowledge graphs. Although relation extraction based on either language information or knowledge graphs alone is a widely researched area, existing techniques are unable to realize the benefits arising from the confluence of language and graph information. Further, the applicability of multimodal methods is limited when data are incomplete or a subset of modalities is missing altogether.

\item \textbf{What this paper adds:} We develop a flexible multi-modal approach for disease relation extraction that de-couples language and graphs but trains a joint model to address the challenge of incomplete data modalities. We also built a human-annotated dataset and evaluate our method on it, demonstrating the effectiveness of the approach over both language- and graph-based methods.
\end{itemize}

\section{Methods}\label{sec:methods}

We next detail the REMAP approach and illustrate it in detail for the task of disease relation extraction and classification (Figure~\ref{Fig. 1}). We first describe the notation, proceed with an overview of language and knowledge graph models, and outline the multimodal learning strategy to inject knowledge into extraction tasks. 




\subsection{Preliminaries}\label{sec:preliminaries}

\xhdr{Notation} 
The input to REMAP is a combined dataset $D$ of language and graph information. This dataset consists of language information $D^T = \{B_i\}_{i=1}^{M_0}  = D^T_L \cup D_{U}^T = \{B_i\}_{i=1}^{M} \cup \{B_i\}_{j=M+1}^{M_0}$ given as  $M_0$ bags of sentences and graph information $D^G = \{(s_i, r_i, o_i)\}_{i=1}^M \bigcup \{(s_i, r_i, o_i)\}_{i=M+1}^N$ given as $N$ triplets $(s_i, r_i, o_i)$ encoding the relationship between $s_i$ and $o_i$ as $r_i$. For example, $s_i = \textrm{``Hypobetalipoproteinemia''}$, $o_i = \textrm{``fatty liver''}$ and $r_i = \textrm{``May Cause''}$ would indicate the fatty liver would be a possible symptom of hypobetalipoproteinemia. We assume that $M$ bags of sentences in $D^T$ overlap with the triplets from existing KG such that each sentence of the $i$th sentence bag contain $(s_i, r_i)$. The remaining $N-M$ triplets in $D^G$ cannot be mapped to sentences in $D^T$, and $M_0-M$ sentences contain entity pairs that do not belong to existing KG. 
We represent the $i$-th sentence bag as  $B_i=\{(t_{ij}, I^s_{ij}, I^o_{ij})\}_{j=1}^{l_i}$, where $l_i$ is the number of sentences in bag $B_i$, $t_{ij}$ is the tokenized sequence of $j$-th sentence in $B_i$. Here, the tokenized sequence is a combination of the mentions of subject and object entities, entity markers, the document title, and the article structure. Marker tokens are added to each entity's head and tail position to denote entity type information. Last, $I^s_{ij}$ and $I^o_{ij}$ are start indices of entity markers for subject and object entities, respectively.


\xhdr{Heterogeneous graph attention network} 
Heterogeneous graph attention network (HAN)~\cite{wang2019heterogeneous} is a graph neural network to embed a KG by leveraging meta paths. A meta path is a sequence of node types and relation types~\cite{sun2011pathsim}. For example, in a disease KG, ``Disease'' $\rightarrow$ ``May Cause'' $\rightarrow$ ``Disease'' $\rightarrow$ ``Differential Diagnosis'' $\rightarrow$ ``Disease'' is a meta path. Node $u_i$ is connected to node $u_j$ via a meta path $\Phi$ if $u_i$ and $u_j$ are the head and tail nodes, respectively, of this meta path. Each node $u_i$ has an initial node embedding $\mathbf{h}_i^{\textrm{init}}$ and belongs to a type $\phi_i$, e.g., $\phi_i = \textrm{``disease concept''}$. Graph attention network specified a parameterized deep transformation function $f_{\textrm{HAN}}$ that maps nodes to condensed data summaries, i.e.,  embeddings, in a node-type specific manner as: $\mathbf{h}^{\prime}_i = f_{\textrm{HAN}}^{\phi_i}(\mathbf{h}_i^{\textrm{init}})$.

We denote all nodes adjacent to $u_i$ via a meta-path $\Phi$ as $u_j \in N_i^{\Phi}$ and node-level attention mechanism provides information on how strongly $f_{\textrm{HAN}}$ attends to $u_i$'s each adjacent node $u_j$ when generating the embedding for $u_i$. In particular, the importance of $u_j$ for $u_i$ in meta path $\Phi$ is defined as: 
\begin{equation}\label{equation:5}
a_{ij}^\Phi=\frac{\textrm{exp}(\sigma(\mathbf{a}_\Phi^T\cdot[\mathbf{h}_i^\prime||\mathbf{h}_j^\prime]))}{\sum_{k\in N_i^\Phi}\textrm{exp}(\sigma(\mathbf{a}_\Phi^T\cdot[\mathbf{h}_i^\prime||\mathbf{h}_k^\prime]))},
\end{equation}
where $\sigma$ is the sigmoid activation, $||$ indicates concatenation, and $\mathbf{a}_\Phi$ is a trainable vector. To promote stable attention, HAN uses multiple, i.e., $K$, heads and concatenates $K$ vectors after node level attention to produce the final node embedding for node $u_i$: 
\begin{equation}\label{equation:6}
\mathbf{z}_i^{\Phi} = ||_{k=1}^K \sigma(\sum_{j\in N_i^{\Phi}}a_{ij}^{\Phi}\cdot \mathbf{h}^\prime_j),
\end{equation} 
%
Given user-defined meta paths $\Phi_1, \ldots, \Phi_P$, HAN uses the above specified node-level attention to produce node embeddings $\mathbf{Z}_{\Phi_1}, \ldots, \mathbf{Z}_{\Phi_P}$. Finally, HAN uses semantic-level attention to combine meta path-specific node embeddings as:
\begin{equation}\label{equation:7}
\beta_{\Phi_p} = \frac{\textrm{exp}(\frac{1}{|E|}\sum_{i \in E}\mathbf{q}^T\cdot \textrm{tanh}(\mathbf{W}\cdot \mathbf{z}_i^{\Phi_p}+b))}{\sum_{p=1}^P \textrm{exp}(\frac{1}{|E|}\sum_{i \in E}\mathbf{q}^T\cdot \textrm{tanh}(\mathbf{W}\cdot \mathbf{z}_i^{\Phi_p}+b))},
\end{equation}
\begin{equation}\label{equation:8}
\mathbf{Z} = \sum_{p=1}^P\beta_{\Phi_p} \cdot \mathbf{Z}_{\Phi_p},
\end{equation}
where $\beta_{\Phi_p}$ represents the importance of meta path $\Phi_p$ towards final node embeddings $\mathbf{Z}$, and $\mathbf{q}^T$, $\mathbf{W}$, and $b$ are trainable parameters. The final outputs are node embeddings $\mathbf{Z}$, representing compact vector summaries of knowledge associated with each node in the KG.

\xhdr{Translation- and tensor-based embeddings}
TransE~\cite{bordes2013translating} and TuckER~\cite{balavzevic2019tucker} decode an optimized set of embeddings into the probability estimate of relationship $r_i$ existing between entities $s_i$ and $o_i$. This is achieved by a scoring function (SF) that either translates the embeddings in TransE as:
$
\textrm{SF}_{\textrm{Tr}}(\mathbf{h}_{s_i}, \mathbf{h}_{o_i}, \mathbf{h}_r) = \sigma(||\mathbf{h}_{s_i} + \mathbf{h}_r - \mathbf{h}_{o_i}||^2_2)
$
or factorizes the embeddings in TuckER as:
$
\textrm{SF}_{\textrm{Tu}}(\mathbf{h}_{s_i}, \mathbf{h}_{o_i}, \mathbf{h}_r, \mathbf{W}) = \sigma(\mathbf{W} \times_1 \mathbf{h}_{s_i} \times_2 \mathbf{h}_r \times_3 \mathbf{h}_{o_i})
$, where $\mathbf{h}_{s_i}$, $\mathbf{h}_{o_i}$, and $\mathbf{h}_r$ represent entity and relation embeddings, $\sigma$ denotes the sigmoid function, $\mathbf{W} \in \mathbb{R}^{d_{s_i}\times d_r\times d_{o_i}}$ is a trainable tensor, and $\times_d$, $d=1,2,3$, indicates tensor multiplication along dimension $d$. 
 
\subsection{Text and knowledge graph encoders}    \label{sec:textencoder}

\xhdr{Embedding disease-associated sentences}
%
We start by tokenizing entities in sentences and proceed with an overview of the language encoder. Entity tokens identify the position and type of entities in a sentence~\cite{MTB,zhong2020frustratingly}.  
Specifically, tokens $\textrm{<S-type>}$ and $\textrm{<S-type/>}$, and $\textrm{<O-type>}$ and $\textrm{<O-type/>}$ are used to denote the start and end of subject (S) and object (O) entities, respectively. The entity marker tokens are type-related, meaning that entities of different types (e.g., disease concepts, medications) get different tokens. 
%
%
%
This procedure produces bags of tokenized sentences $D^T$ that we encode into entity embeddings using a language encoder. We use SciBERT encoder~\cite{scibert} with the SciVocab  vocabulary, which is a BERT language model optimized for scientific text with improved efficiency in biomedical domains than BioBERT or BERT model alone~\cite{zhong2020frustratingly}. 
Tokenized sequences in a sentence bag $B_i$ are fed into the language model to produce a set of sequence outputs $\Hbb_i = \{\mathbf{H}_i\}_{i=1}^{l_i}, i=1,2,\ldots,l_i$:
\begin{equation}\label{equation:1}
\Hbb_i^{[m]}  = \textrm{SciBERT}(B_i; \Hbb_i^{[m-1]}),
\end{equation}
where $\Hbb_i = [\bH_{i1}, ..., \bH_{il_i}]$, $l_i$ is the number of sentences in $B_i$, $\mathbf{H}_i \in \mathbb{R}^{d_l \times d_{hs}}$. We then aggregate representations of subject entities $s_i$ across all sentences in bag $B_i$ as: $\mathbf{H}_{s_i} = ||_{m=1}^{l_i}||_{k \in I^s_{ij}}\mathbf{H}_{mk}$ 
%
%
%
and use self-attention to obtain the final language-based embedding $\mathbf{h}^T_{s_i}$ for subject entity $s_i$ as:
\begin{equation}\label{equation:3}
\mathbf{h}^T_{s_i} =\mathbf{H}_{s_i} \cdot \textrm{softmax}(\boldsymbol{\omega} \cdot \textrm{tanh}(\mathbf{H}_{s_i})),
\end{equation}
where $\mathbf{h}^T_{s_i} \in \mathbb{R}^{d_{hs}}$ is the embedding of $s_i$ and $\boldsymbol {\omega}$ is a trainable vector. Embeddings of object entities (i.e., $\mathbf{h}^T_{o_i}$ for object entity $o_i$) are generated analogously by the language encoder. Self-attention is needed because specific sentences in a bag may not contain disease-disease relationships, and the attention mechanism allows the model to down-weight those uninformative sentences when generating embeddings.


\xhdr{Embedding disease-disease knowledge relationships}
We use a heterogeneous graph attention encoder~\cite{wang2019heterogeneous} to derive embeddings for nodes in the disease knowledge graph. The encoder produces embeddings for every subject entity $\mathbf{h}^G_{s_i}$ and every object entity $\mathbf{h}^G_{o_i}$ that have corresponding nodes in the KG as follows:
\begin{equation}\label{equation:20}
\mathbf{h}^G_{s_i} = f_{\textrm{HAN}}(D^G, \mathbf{H}^{\textrm{init}}, s_i),\quad
\mathbf{h}^G_{o_i} = f_{\textrm{HAN}}(D^G, \mathbf{H}^{\textrm{init}}, o_i),
\end{equation}
where $f_{\textrm{HAN}}$ transformation is given in Eqs.~(\ref{equation:5})-(\ref{equation:8}) and $\mathbf{H}^{\textrm{init}}$ denotes the matrix of initial embeddings.

\xhdr{Scoring disease-disease relationships} 
Taking language-based embeddings, $\mathbf{h}^T_{s_i}, \mathbf{h}^T_{o_i} $, and graph-based embeddings, $\mathbf{h}^G_{s_i}, \mathbf{h}^G_{o_i}$, for diseases that appear in either language or graph dataset, REMAP scores triplets $(s_i, r_k, o_i)$ as candidate disease-disease relationships. Specifically, to estimate the probability that diseases $s_i$ and $o_i$ are associated through a relation of type $r_k$ (e.g., $r_k = \textrm{``May Cause''}$), REMAP calculates scores $p^T$ and $p^G$ representing the amount of evidence in the combined language-graph dataset that supports the disease-disease relationship:
\begin{align}\label{equation:9}
p^T(r_{ik}=1|s_i, o_i) = \textrm{SF}(\mathbf{h}^T_{s_i}, \mathbf{h}^T_{o_i}, \mathbf{h}_r),\;k=1,2,\ldots,K,\\
p^G(r_{ik}=1|s_i, o_i) = \textrm{SF}(\mathbf{h}^G_{s_i}, \mathbf{h}^G_{o_i}, \mathbf{h}_r),\;k=1,2,\ldots,K,
\end{align}
where $\textrm{SF}$ is the scoring function, and $K$ denotes the number of relation types. We consider three scoring functions, including the linear scoring function:
$
\textrm{SF}_{\textrm{Li}}(\mathbf{h}_{s_i}, \mathbf{h}_{o_i}, \mathbf{h}_{r_k}) = \sigma(\mathbf{W}_k(\mathbf{h}_{s_i} + \mathbf{h}_{o_i})+b_k)
$,
the TransE scoring function:
$
\textrm{SF}_{\textrm{Tr}}(\mathbf{h}_{s_i}, \mathbf{h}_{o_i}, \mathbf{h}_{r_k}) = \sigma(||\mathbf{h}_{s_i} + \mathbf{h}_{r_k} - \mathbf{h}_{o_i}||^2_2)
$,
and the TuckER scoring function:
$
\textrm{SF}_{\textrm{Tu}}(\mathbf{h}_{s_i}, \mathbf{h}_{o_i}, \mathbf{h}_{r_k}, \mathbf{W}) = \sigma(\mathbf{W} \times_1 \mathbf{h}_{s_i} \times_2 \mathbf{h}_{r_k} \times_3 \mathbf{h}_{o_i}),
$
where separate kernels $\mathbf{W}^T$ and $\mathbf{W}^G$ are used for language and graph in the TuckER decomposition, and $\mathbf{h}_{r_k}$ denotes the encoding of relation type $r_k$ in TransE that is shared across both modalities (Section~\ref{sec:preliminaries}).

\subsection{Co-training text and graph encoders}\label{sec:jointlearning}

We proceed to describe the procedure for co-training text and graph encoders. 
%
From last section, we obtain relationship estimates based on evidence provided by text information, $p^T(r_{ik}=1|s_i, o_i)$, and graph information, $p^G(r_{ik}=1|s_i, o_i)$, for every triplet $(s_i, r_i, o_i)$. We use the binary cross entropy to optimize those estimates in each data type as:
\begin{equation}\label{equation:14}
L^T =\sum_{i=1}^M\sum_{k=1}^K r_{ik} \log(p^T(r_{ik}=1|s_i, o_i)) + (1 - r_{ik}) \log(1-p^T(r_{ik}=1|s_i, o_i)),
\end{equation}
\begin{equation}\label{equation:15}
L^G =\sum_{i=1}^M\sum_{k=1}^K r_{ik} \log(p^G(r_{ik}=1|s_i, o_i)) + (1 - r_{ik}) \log(1-p^G(r_{ik}=1|s_i, o_i)),
\end{equation}
and can combine language-based loss $L^T$ and graph-based loss $L^G$ as: $L_{\textrm{REMAP}} = L^T + L^G$. This loss function is motivated by the principle of knowledge distillation~\cite{lan2018knowledge} to enhance multimodal interaction and improve classification performance. Using probabilities $p^T(r_{ik}=1|s_i, o_i)$ from the language encoder, we normalize them across $K$ relation types to obtain distribution $\mathbf{p}_t(s_i, o_i)$ as: 
\begin{equation}\label{equation:17}
\mathbf{p}^T(s_i, o_i) = \{\frac{e^{p^T(r_{ik}=1|s_i, o_i)}}{\sum_{m=1}^K e^{p^T(r_{im}=1|s_i, o_i)}}\}_{k=1}^K.
\end{equation}
In the same manner, we calculate a graph-based distribution $\mathbf{p}^G(s_i, o_i)$ using softmax normalization. 

Specifically, we develop two REMAP variants, REMAP-M and REMAP-B, based on how language-based and graph-based losses are combined into a multimodal objective. In REMAP-M, both losses are aligned by shrinking the distance between distributions $\mathbf{p}^T$ and $\mathbf{p}^G$ using the Kullback-Leibler (KL) divergence:
\begin{equation}\label{equation:18}
D_{\textrm{KL}}(\mathbf{p}^T, \mathbf{p}^G) = \sum_{i=1}^M\sum_{k=1}^K p^G(s_i, o_i)_k \log(\frac{p^G(s_i, o_i)_k}{p^T(s_i, o_i)_k}),
\end{equation}
where we measure the misalignment between language and graph models as follows:
\begin{equation}\label{equation:19}
L_{\textrm{REMAP-M}} = L_{\textrm{REMAP}} + \lambda_{M}(D_{\textrm{KL}}(\mathbf{p}^T, \mathbf{p}^G) + D_{\textrm{KL}}(\mathbf{p}^G, \mathbf{p}^T)).
\end{equation}
Instead of measuring how distribution $\mathbf{p}^T$ is different from $\mathbf{p}^G$, REMAP-B selects the strongest logit across data types using an ensemble distillation strategy~\cite{guo2020online}. Specifically, REMAP-B uses the highest predicted score $p^B$ across both language and graph models to derive final predictions:
\begin{align}
    p^B(r_{ik}=1|s_i, o_i)=
    \left\{
    \begin{array}{cc}
    p^T(r_{ik}=1|s_i, o_i), &\;\;p^T(r_{ik}=1|s_i, o_i) \le p^G(r_{ik}=1|s_i, o_i)  \;\; \textrm{and} \;\; r_{ik}=0\\
    p^T(r_{ik}=1|s_i, o_i), & \;\;p^T(r_{ik}=1|s_i, o_i) \ge p^G(r_{ik}=1|s_i, o_i)  \;\; \textrm{and} \;\; r_{ik}=1\\
    p^G(r_{ik}=1|s_i, o_i), & \;\;p^T(r_{ik}=1|s_i, o_i) \ge p^G(r_{ik}=1|s_i, o_i)  \;\; \textrm{and} \;\; r_{ik}=0\\
    p^G(r_{ik}=1|s_i, o_i), & \;\;p^T(r_{ik}=1|s_i, o_i) \le p^G(r_{ik}=1|s_i, o_i)  \;\; \textrm{and} \;\; r_{ik}=1\\
    \end{array}
    \right.
\end{align}
that are softmax-normalized across $K$ relation types:
\begin{equation}
\textbf{p}^B(s_i, o_i) = \{\frac{e^{p^B(r_{ik}=1|s_i, o_i)}}{\sum_{m=1}^K e^{p^B(r_{im}=1|s_i, o_i)}}\}_{k=1}^K.
\end{equation}
Finally, to minimize the discrepancy between predicted and known disease-disease relationships, REMAP-B uses minimizes the cross-entropy function: 
\begin{equation}
L^B = \sum_{i=1}^M\sum_{k=1}^K r_{ik} \log(p^B(r_{ik}=1|s_i, o_i)) + (1 - r_{ik}) \log(1-p^B(r_{ik}=1|s_i, o_i)),
\end{equation}
and co-trains the multimodal encoder by promoting predictions that are aligned by both encoders:
\begin{equation}
L_{\textrm{REMAP-B}} = L_{\textrm{REMAP}} + \lambda_{B} [L^{B} + D_{\textrm{KL}}(\mathbf{p}^B,\mathbf{p}^T) + D_{\textrm{KL}}(\mathbf{p}^B,\mathbf{p}^G)].
\end{equation}
The outline of the complete REMAP-B algorithm is shown in Algorithm~\ref{algo:REMAP}. 




\begin{algorithm}[t]
	\caption{\textbf{REMAP, multimodal learning on graphs for disease relation extraction and classification.} Shown is the outline of REMAP-B (Section~\ref{sec:jointlearning}).}
	\label{algo:REMAP}
	{\small\LinesNumbered
	\KwIn{Bag of sentences $\{B_i\}_{i=1}^n$, 
	Knowledge graph $D^G = \{(s_i, r_i, o_i)\}$, Initial node embeddings $\mathbf{H}^{\textrm{init}}$, Scoring function $\textrm{SF}$, Regularization strength $\lambda_B$, Relation type vector pretrained on language dataset $\mathbf{r}^T$, Relation type vector pretrained on graph dataset $\mathbf{r}^G$}
	\KwOut{Model parameters $\Theta$ of language-based and graph-based encoders}
	Initialize model parameters $\Theta$\\
	Initialize relation representation as $\mathbf{r}=\frac{\mathbf{r}^T+\mathbf{r}^G}{2}$\\
	\For{epoch=1:n\_epochs}{
	\For{i=1:n\_bags}{
	    /* Encode language and the calculation logits, see Section~\ref{sec:textencoder} */ \\ 
	    $\mathbf{H} = \textrm{SciBERT}(B_i)$ \\
	    $\mathbf{H}_{s_i} = ||_{m=1}^{l_i}||_{k \in I^s_{ij}}\mathbf{H}_{mk}$ \\
	    $\mathbf{H}_{o_i} = ||_{m=1}^{l_i}||_{k \in I^o_{ij}}\mathbf{H}_{mk}$ \\
	    $\mathbf{h}^T_{s_i} =\mathbf{H}_{s_i} \cdot \textrm{softmax}(\boldsymbol {\omega} \cdot \textrm{tanh}(\mathbf{H}_{s_i}))$ \\
	    $\mathbf{h}^T_{o_i} =\mathbf{H}_{o_i} \cdot \textrm{softmax}(\boldsymbol {\omega} \cdot \textrm{tanh}(\mathbf{H}_{o_i}))$ \\
	    $\mathbf{p}^T(s_i, o_i) = \textrm{SF}(\mathbf{h}^T_{s_i}, \mathbf{h}^T_{o_i}, \mathbf{r})$ \\
	    $L^T = \textrm{BinaryCrossEntropyLoss}(\mathbf{p}^T(s_i, o_i), \mathbf{r}(s_i, o_i))$ \\
	    
	    /* Encode graph and calculate graph logits, see Section~\ref{sec:textencoder} */
	    
	    $\mathbf{h}^G_{s_i}=\textrm{HAN}(G, \mathbf{H}^{init}, s_i)$ \\
	    $\mathbf{h}^G_{o_i}=\textrm{HAN}(G, \mathbf{H}^{init}, o_i)$ \\
	    $p^G(s_i, o_i) = \textrm{SF}(\mathbf{h}^G_{s_i}, \mathbf{h}^G_{o_i}, \mathbf{r})$  \\
	    $L^G = \textrm{BinaryCrossEntropyLoss}(\mathbf{p}^G(s_i, o_i), \mathbf{r}(s_i, o_i))$ \\
	    
	    /* Find the best logit and calculate alignment penalty loss, see Section~\ref{sec:jointlearning} */
	    
	    $\mathbf{p}^B(s_i, o_i) = \textrm{CalculateBestLogic}(\mathbf{p}^T(s_i, o_i), \mathbf{p}^G(s_i, o_i))$ \\
	    $\mathbf{p}^T(s_i, o_i) = \textrm{softmax}(\mathbf{p}^T(s_i, o_i))$\\
	    $\mathbf{p}^G(s_i, o_i) = \textrm{softmax}(\mathbf{p}^G(s_i, o_i))$\\
	    $\mathbf{p}^B(s_i, o_i) = \textrm{softmax}(\mathbf{p}^B(s_i, o_i))$\\
	    $L^B = \sum_{i=1}^M\sum_{k=1}^K r_{ik} \log(p^B(r_{ik}=1|s_i, o_i)) + (1 - r_{ik}) \log(1-p^B(r_{ik}=1|s_i, o_i)) $\\
	    $L_{\textrm{REMAP-B}} = L^T + L^G + \lambda_{B} [L^{B} + D_{\textrm{KL}}(\mathbf{p}^B, \mathbf{p}^T) + D_{\textrm{KL}}(\mathbf{p}^B, \mathbf{p}^G)]  $ \\
	    $\Theta  \leftarrow \textrm{Update}(\Theta, L_{\textrm{REMAP-B}})$\\
	}}}
\end{algorithm}

\section{Experiments}\label{sec:data-setup}
We proceed with the description of datasets (Section \ref{sec:Dataset}), followed by implementation details of REMAP approach (Section \ref{sec:TrainImplementation}) and the outline of experimental setup (Section \ref{sec:ExperimentalSetup}).

\subsection{Datasets}\label{sec:Dataset}

Datasets used in this study are multimodal and originate from diverse sources that we integrated and harmonized, as outlined below. In particular, we compiled a large disease-disease KG with text descriptions retrieved from medical data repositories using distant supervision following the data collection and preprocessing strategy described in Lin \etal~\cite{lin2020highthroughput}. Details on data preprocessing and feature engineering are described in \ref{sec:data_preprocessing}. Further, we utilize a large EHR dataset previously validated in Beam \etal~\cite{beam2019clinical} and a novel human-annotated dataset.


\xhdr{Disease knowledge graph} 
We construct a disease-disease KG from Diseases database~\cite{pletscher2015diseases} and MedScape~\cite{frishauf2005medscape} repository that unifies evidence on disease-disease associations based on text mining, manually curated literature, cancer mutation data, and genome-wide association studies. We construct a KG between diseases following the approach outlined in Lin \etal~\cite{lin2020highthroughput}.
This KG contains 9,182 disease concepts, which are assigned concept unique identities (CUI) in the Unified Medical Language System (UMLS), and three types of relationships between disease concepts, which are `may cause' (MC), `may be caused by' (MBC), and `differential diagnosis' (DDx). Other relations in the KG are denoted as `not available' relations (NA). The MBC relation type is the reverse relation of the MC relation type, while DDx is a symmetric relation between disease concepts. Dataset statistics are in \cite{lin2020highthroughput} and Table~\ref{tab:test}.

\xhdr{Language dataset} 
We use a text corpus taken from Lin \etal~\cite{lin2020highthroughput}. This corpus is built from medicine-related articles, including 42 million Pubmed abstracts\footnote{Downloaded from The PubMed Baseline Repository provided by the National Library of Medicine, January 2021}, web pages collected via web crawler from 27 thousand pages on Uptodate and Medscape eMedicine, 10 thousand Wikipedia articles with medical titles, and the main text of four textbooks\footnote{\emph{Harrison's Principles of Internal Medicine 20th Edition, Kelley's Textbook of Internal Medicine 4th Edition, Sabiston Textbook of Surgery: The Biological Basis of Modern Surgical Practice 20th Edition, and Kumar and Clark's Clinical Medicine 7th Edition}}. The corpus is segmented into sentences, which includes 237,119,572 sentences. We first employ forward maximum matching to identify all UMLS disease concepts in the corpus. Then, we link triplets in our disease knowledge graph to sentences if subject and object entities are both in the sentences. This method allows 1,466,065 sentences from these articles to be aligned to disease-disease edges in the disease knowledge graph. In the end, we group sentences by triplets. For example, the triplet \emph{(Hypobetalipoproteinemia, may cause Fatty liver)} is aligned to a bag of sentences containing 4 sentences.  More statistical details are shown in Table~\ref{tab:test}. 

\xhdr{Electronic health record dataset}
We use two types of information from electronic health records (EHRs), both taken from Beam \etal~\cite{beam2019clinical}. In particular, with a nationwide US health insurance plan with 60 million members over the period of 2008-2015, a dataset of concept co-occurrences from 20 million notes at Stanford, and an open access collection of 1.7 million full-text journal articles obtained from PubMed Central, Beam \etal~\cite{beam2019clinical} created a dataset of disease concepts identified with SNOMED-CT that appear together in the same note and used the SVD to decompose the resulting co-occurrence matrix and produce 500 dimension embedding vectors for disease concepts. We use the information on co-occurring disease concepts to guide the sampling of negative node pairs when training the disease knowledge graph. Further, we use the 500 dimension embedding vectors from \cite{beam2019clinical} to initialize embeddings in REMAP's graph neural network. We use the average of all concept embedding as their initial embeddings for out-of-dictionary disease concepts.


\xhdr{Human annotated dataset}
Relation triplets retrieved from databases may still have few incorrect data. To build a gold test set for a robust evaluation of our model, we randomly selected 500 disease triplets from our disease knowledge graph. We omit this subgraph from the knowledge graph used for model training and create an annotated dataset with it. For that, we recruited three senior physicians from Peking Union Medical College Hospital. We require them independently assign candidate relations \emph{(May cause, May be caused, DDx and Not Available)} to these entity pairs without showing them relations collected from the database. We find annotation experts only disagreed on labels for  14 disease pairs (\ie, 2.8\% of the total number of disease pairs), and we resolve these disagreements through consensus voting. The human-annotated dataset is used for model comparison and performance evaluation.

\begin{table}[t]
\small
 \caption{\label{tab:test} \textbf{Overview of the disease knowledge graph and the language dataset.} Shown are statistics for the following relation types: Not Available (NA), Differential Diagnosis (DDx), May Cause (MC), and May Be Caused by (MBC). Total denotes the total number of all triplets (see Figure~\ref{Fig2}).}
 \centering
 \begin{tabular}{lll|cccccc}
  \toprule
  \multicolumn{3}{c}{Dataset} & Total & NA & DDx & MC & MBC & Entities \\
  \midrule
  Unaligned & - & - & 96,913 & 30,546 & 20,657 & 23,411 & 22,298 & 9,182 \\
  \hline
  \multirow{6}{*}{Aligned} & \multirow{2}{*}{Train} & Triplet & 31,037 & 15,670 & 7,262 & 4,358 & 3,747 & \multirow{2}{*}{7,794} \\
  & & Language & 1,244,874 & 799,194 & 208,921 & 123,735 & 113,024 & \\
  \cline{2-9}
  & \multirow{2}{*}{Validation} & Triplet & 7,754 & 3,918 & 1,821 & 1,065 & 950 & \multirow{2}{*}{4,433} \\
  & & Language & 206,179 & 68,934 & 60,165 & 43,706 & 33,474 & \\
  \cline{2-9}
  & \multirow{2}{*}{Annotated} & Triplet & 499 & 8 & 210 & 159 & 122 & \multirow{2}{*}{733} \\
  & & Language & 15,012 & 96 & 4,980 & 6,699 & 3,237 & \\
  \bottomrule
 \end{tabular}
\end{table}

\subsection{Training REMAP models}\label{sec:TrainImplementation}

Next, we outline the training details of REMAP models, including negative sampling, pre-training strategy for language and graph models, and the multimodal learning approach.

\xhdr{Negative sampling} 
We construct negative samples by sampling disease pairs whose co-occurrence in the EHR co-occurrence matrix~\cite{beam2019clinical} is lower than a pre-defined threshold. In particular, we expect that two unrelated diseases rarely appear together in EHRs, meaning that the corresponding values in the co-occurrence matrix are low. Thus, such disease pairs represent suitable negative samples to train models for classifying disease-disease relations. 


\xhdr{Pre-training a language model} 
The text-based model comprises the text encoder and the TuckER module for disease-disease relation prediction. We denote the relation embeddings as $\mathbf{r}^T$, and the loss function as $L^T$ (Section~\ref{sec:jointlearning}). In particular, we use SciBERT tokenizer and SciBERT-SciVocab-uncased model~\cite{wolf2019huggingface}. The entity markers are added to the SciVocab vocabulary, and their embeddings are initialized with uniform distribution. We set the maximum number of sentences in a bag to $l_{m(max)}$. If the bag size is greater than $l_{m(max)}$, then $l_{m(max)}$ sentences are selected uniformly at random for training. Further details on hyper-parameters are in Table~\ref{tab:param}.


\xhdr{Pre-training a graph neural network model} 
The graph-based model comprises the heterogeneous attention network encoder and the TuckER module for disease-disease relation prediction. The relation embeddings produced by the TuckER module are denoted as $\mathbf{r}^G$. In the pre-training phase, the model is optimized for the loss function is $L^G$ (Section~\ref{sec:jointlearning}). The initial embeddings for nodes in the disease knowledge graph are concept unique identifier (CUI) representations derived from the SVD decomposition of the EHR co-occurrence matrix~\cite{beam2019clinical}. Further details on hyper-parameters are in Table~\ref{tab:param}.


\xhdr{Cross-modal learning}
After data type-specific pre-training is completed, the text and graph models are fused in cross-modal learning. To this end, the shared relation vector $\mathbf{r}$ is initialized as: $\mathbf{r}=(\mathbf{r}^T+\mathbf{r}^G)/2$. We consider two REMAP variants, namely REMAP-M and REMAP-B, optimized for different loss functions (Section~\ref{sec:jointlearning}). Details on hyper-parameter selection are in Table~\ref{tab:param}.

\subsection{Experimental setup}\label{sec:ExperimentalSetup}

Next we overview baseline methods and performance metrics.

\xhdr{Baseline methods} We consider 9 baseline methods divided into two groups: 5 methods for link prediction on KGs and 5 methods for relation extraction in text. 

For graph-based baselines, we take both disease and relation embeddings as parameters. We randomly initialize the embeddings and train the knowledge graph models or graph neural networks. Graph-based baselines are trained on the disease KG and include the following:
\begin{itemize}
    \item \textbf{TransE}~\cite{bordes2013translating} embeds diseases and relations by translating embedding vectors in the learned embedding space. Given embeddings of a triplet $(\mathbf{h},\mathbf{r},\mathbf{t})$, the score is calculated as $s=||\mathbf{h}+\mathbf{r}-\mathbf{t}||_2^2$. We train our TransE model using negative sampling and $L2$ penalty as recommended by the authors. The negative samples are constructed by randomly replacing objects $\mathbf{t}$ given subject $\mathbf{h}$ and relation $\mathbf{r}$. We begin with random embeddings for diseases and relations and then optimize them with the same margin-based ranking criterion and stochastic gradient descent as in the original TransE paper.
    \item \textbf{DistMult}~\cite{DistMult} predicts disease-disease relationships using a bilinear decoder for edge in the KG. Given embeddings of a triplet $(\mathbf{h}, \mathbf{r}, \mathbf{t})$, the score is calculated as $s=\mathbf{h}\mathbf{M}_{r}\mathbf{t}$, where $\mathbf{M}_r$ is a trainable diagonal matrix for relation $r$. The disease and relation embeddings are parameters in this baseline. We initialize it with random embeddings and optimize them with margin-based ranking loss and batch stochastic gradient descent. The negative samples are constructed by corrupting the subject or object in a relation triplet.
    \item \textbf{ComplEx}~\cite{ComplEx} predicts disease-disease relationships by carrying out a matrix factorization using complex-valued embeddings. Denoting the complex-valued embeddings of a relation triplet as $\mathbf{h}_s$, $\mathbf{w}_r$, and $\mathbf{h}_o$. ComplEx assumes $P(Y=1)=\sigma(\phi(\mathbf{h}_s,\mathbf{w}_r,\mathbf{h}_o))$, where $Y=1$ represents the triplet $(h, r, t)$ holds and $\sigma$ denotes the sigmoid function. The score function $\phi(\cdot)$ is calculated as $s=Re(<\mathbf{h},\mathbf{r},\mathbf{t}>)$. The product $<\cdot>$ is a Hermitian product, and relation embedding $\mathbf{r}$ is a complex-valued vector. We take random embeddings as initial embeddings for both real and imaginary parts. The training object is minimizing the negative log-likelihood of this logistic model with $L_2$ penalty on the parameters of disease and relation embeddings.
    \item \textbf{RGCN}~\cite{schlichtkrull2017modeling} predicts disease-disease relationships using relational graph convolutional network (RGCN). Following the authors' recommendations, we use a 2-layer RGCN to embed the KG and DistMult decoder for link prediction. We train the RGCN model using the cross-entropy loss on four relations and the Adam optimizer. The RGCN uses the sigmoid activation function and has a 0.4 dropout rate for self-loops and 0.2 for other edges.
    \item \textbf{TuckER}~\cite{balazevic2019tucker} is a KG embedding method that uses Tucker tensor decomposition. Given the embeddings of a triplet $(\mathbf{h}, \mathbf{r}, \mathbf{t})$, the score is calculated as $s=\mathbf{W} \times_1 \mathbf{h} \times_2 \mathbf{r} \times_3 \mathbf{t}$, where $\mathbf{W}$ is a trainable matrix and $\times_i$ denotes tensor multiplication on dimension $i$. 
\end{itemize}

Text-based methods are trained on the language dataset following Lin \etal~\cite{lin2020highthroughput} strategy. 
We consider models with bag-of-words (BoW) engineered features, convolutional neural networks, and pre-trained language models. Sentence-level baselines, including RF, TextCNN, BiGRU, and PubmedBERT, use majority voting on sentence-level predictions to produce final predictions for input examples that are longer than one sentence. We consider the following text-based baselines:
\begin{itemize}
    \item \textbf{Random Forest (RF)} uses scaled BoW features with 100 decision trees and the Gini criterion as the classifier. Punctuations and English stop words are removed from sentences \cite{manning2014stanford}. We extract 40,000 most frequently occurring N-grams and calculate the TF-IDF to produce sentence-level features to RF. The N-gram TF-IDF transformation is a widely-used and effective feature construction procedure.
    \item \textbf{TextCNN}~\cite{textcnn} is a convolutional neural network for text, a useful deep learning algorithm for sentence classification tasks, such as sentiment analysis and question classification. The TextCNN we take as a baseline has 64 feature maps for each size and 8 different sizes of feature maps whose lengths of filter windows range from 2 to 10. We use ReLU as the activation function, 1 dimension max pooling after convolutional layers, and a dropout layer with a 0.25 dropout rate. We use skip-gram embeddings \cite{mikolov2013distributed} to initialize the TextCNN model and train it using cross-entropy loss and Adam optimizer.
    \item \textbf{BiGRU} \cite{cho2014properties} is a recurrent neural network initialized in the same way as TextCNN. The encoder is a 1-layer BiGRU whose output hidden states are aggregated into sentence embedding using a one-headed self-attention. These sentence embeddings are fed into a linear layer and provide logits for relation classification. The dimension of input word embeddings is 200, and the dimension of BiGRU hidden states is 100. We also add dropout layers with a 0.5 dropout rate after BiGRU and the linear layer. The sentence-level predictions given by this model will generate an instance-level classification result by majority voting. 
    \item \textbf{BiGRU+Attention} \cite{cho2014properties} is the same as the BiGRU baseline with an added instance-level attention. Instead of voting, this approach uses one-head instance-level attention to aggregate latent vectors across all sentences in an instance. This provides an instance embedding for each sentence bag. Then we use a linear layer to provide logits of relation classification from instance embedding. This baseline is trained with cross-entropy loss on four relations, including Not Available (NA). We use the same hyper-parameters for this baseline as for BiGRU.
    \item \textbf{PubmedBERT}~\cite{gu2021domain} is a domain-specific pre-trained language model trained on the corpus of abstracts from Pubmed with mask language modeling and next sentence prediction as pretext tasks. We take the checkpoint of PubmedBERT and fine tune the model for disease relation extraction. We use the output embedding of the [CLS] token as sentence embedding. And then sentence embedding in a bag is aggregated with instance-level attention. The fine tuning employs the Adam optimizer and a linear scheduler to adjust the learning rate. The loss function is also cross-entropy loss on four different relations, including Not Available (NA).
\end{itemize}

We use grid search on the validation set to select hyper-parameters for all methods. Table~\ref{tab:param} outlines the hyper-parameter selection in REMAP. 

\xhdr{Performance metrics}
We evaluate predicted disease-disease relations by calculating the accuracy of predicted relations between disease concepts, which is an established approach for benchmarking relation extraction methods~\cite{hu2020open}. Specifically, given a triplet $(s_i, r_i, o_i)$ and a predicted score $p_i \in [0,1]$, relation $r_i$ is predicted to exist between $s_i$ and $o_i$ if the predicted score $p_i \ge \textrm{threshold}_i$, which corresponds to a binary classification task for each relation type. The $\textrm{threshold}_i$ is a relation type-specific value determined such that binary classification performance achieves maximal F1-score. We report classification accuracy, precision, recall, and F1-score for all experiments in this study. 

\subsection{Variants of REMAP approach}
We carry out an ablation study to examine the utility of key REMAP components. We consider the following three components and examine REMAP's performance with and without each. The results of all variants are reported in the ablation study \ref{tab:ablation}.
\begin{itemize}
\item \textbf{REMAP-B without joint learning} In text-only ablations, we use SciBERT to obtain concept embeddings $\mathbf{h}_s^T$ and $\mathbf{h}_o^T$, and combine them to produce a relation embedding $\mathbf{r}_k$, where $\mathbf{r}_k$ is related to the concept embeddings based on text information. Finally, we consider three scoring functions to classify disease-disease relations, and we denote the models as SciBERT (linear), SciBERT (TransE), and SciBERT (TuckER). Similarly, in graph-only ablations, we first use a heterogeneous attention network to obtain graph embeddings $\mathbf{h}_s^G$ and $\mathbf{h}_o^G$ that are combined into prediction by different scoring functions, including HAN (linear), HAN (TransE), and HAN (TuckER).
\item \textbf{REMAP-B without EHR embeddings} REMAP-B uses EHR embeddings~\cite{beam2019clinical} as initial node embedding $\mathbf{H}^{init}$. To examine the utility of EHR embeddings, we design an ablation study that initializes node embeddings using the popular Xavier initialization~\cite{glorot2010understanding} instead of EHR embeddings. Other parts of the model are the same as in REMAP-B.
\item \textbf{REMAP-B without unaligned triplets} Unaligned triplets denote triplets in the disease knowledge graph that do not have the corresponding sentences in the language dataset. To demonstrate how these unaligned triplets influence model performance, we design an ablation study in which we train a REMAP-B model on the reduced disease knowledge graph with unaligned triplets excluded.
\end{itemize}


\section{Results}\label{sec:results}
REMAP is a multimodal language-graph learning approach. We evaluate REMAP's prowess for disease relation extraction when the REMAP model is tasked to identify candidate disease relations in either text (Section~\ref{sec:results-relation-extraction}) or graph-structured (Section~\ref{sec:results-relation-classification}) data. We present ablation study and case studies in the discussion (Section~\ref{sec:discussion}). 


\subsection{Extracting disease relations from text}\label{sec:results-relation-extraction}

We start with results on the human-annotated dataset where each method, while it can be trained on a multimodal text-graph dataset, is tasked to extract disease relations from text alone, meaning that the test set consists of only annotated sentences. Table~\ref{tab:result} shows performance on the annotated set for text-based methods. 
We witness the pre-trained language model, such as PubmedBERT\cite{gu2021domain}, consistently outperforms other neural network baselines and random forest in F1 score, which achieves 78.5 in micro average.  Further, BiGRU+Attention is the best performing baseline method in accuracy, achieving an accuracy of 78.6. We also find that REMAP-B and REMAP-M achieve the best performance across all settings, outperforming baselines significantly. In particular, REMAP models surpass the strongest baseline by 10.0 absolute percentage points (accuracy) and by 7.2 absolute percentage points (F1-score). These results show that multimodal learning can considerably advance disease relation extraction when only one data type is available at test time.

\begin{table}[t]
 \caption{\label{tab:result} \textbf{Results of disease relation extraction on the human-annotated set.} DDx: differential diagnosis, MC: may cause, MBC: may be caused by. The ``micro'' columns denote micro average accuracy or F1-score for DDx, MC, and MBC relation types. Further results are in \ref{sec:further-performance}.}
 \centering
 \setlength{\tabcolsep}{1.5mm}{
 \begin{tabular}{lcl|cccc|cccc}
  \toprule
  \multirow{2}{*}{Modality} &\multicolumn{2}{c}{\multirow{2}{*}{Model}} & \multicolumn{4}{c}{Accuracy} & \multicolumn{4}{c}{F1-score} \\
   &&& micro & DDx & MC & MBC & micro & DDx & MC & MBC  \\
  \midrule
  \multirow{8}{*}{Text} & \multirow{5}{*}{Baselines} 
  &RF & 72.0 & 68.6 & 70.8 & 76.4 & 37.8 & 53.1 & 25.5 & 19.2\\
  &&TextCNN	&76.7	&75.4	&73.0	&81.8&60.9	&67.5	&59.9	&48.6\\
  &&BiGRU	&77.4	&73.0	&77.2	&82.0&62.0	&67.9	&54.0	&59.8\\
  &&BiGRU+attention &78.6 &75.0 &78.2 &82.6&64.6	&67.7	&63.5	&60.6\\
  &&PubmedBERT & 77.7 & 82.4 & 75.5& 82.0 &	78.5 & 75.2 &	74.6 & 81.7\\
  \cline{2-11}
  & \multirow{3}{*}{Ours} &REMAP & 88.2 & 83.6 & 89.0 & 92.0 & 80.9 & 80.7 & 80.0 & 82.6\\
  &&REMAP-M & 88.6 & 84.2 & 89.0 & \textbf{92.8} & 81.5 & 81.6 & 79.6 & \textbf{83.8}\\
  &&REMAP-B & \textbf{88.6} & \textbf{84.4} & \textbf{89.2} & 92.4 &\textbf{81.8} & \textbf{81.9} & \textbf{80.3} & 83.3\\
  \midrule
  \multirow{9}{*}{Graph} & \multirow{5}{*}{Baselines} & TransE\_l2 & 75.1 & 70.7 & 72.7 & 81.8 & 63.2 & 68.0 & 57.0 & 62.2\\
  &&DistMult & 69.8 & 77.5 & 61.3 & 70.5 & 56.1 & 71.0 & 43.4 & 51.5 \\
  &&ComplEx & 79.0 & 75.2 & 77.8 & 84.2 & 65.0 & 69.3 & 56.5 & 66.9 \\
  &&RGCN & 71.8 & 78.6 & 62.5 & 74.3 & 62.2 & 75.1 & 50.8 & 58.6\\
  &&TuckER & 81.5 & 77.6 & 82.3 & 84.7 & 73.7 & 76.2 & 71.7 & 71.9\\
  \cline{2-11}
  &\multirow{3}{*}{Ours} & REMAP & 89.6 & 86.4 & 89.6 & \textbf{92.8} & 83.5 & 84.3 & 81.6 & \textbf{84.2} \\
  &&REMAP-M & 89.3 & 87.0 & 88.4 & 92.6& 83.3 & 85.7 & 78.8 & 83.8 \\
  &&REMAP-B & \textbf{89.8} & \textbf{87.3} & \textbf{89.9} & 92.2& \textbf{84.1} & \textbf{85.8} & \textbf{82.4} & 82.7\\
  \bottomrule
 \end{tabular}}
\end{table}

\subsection{Completing disease KG with novel disease-disease relationships}\label{sec:results-relation-classification}

We proceed with the results of disease relation prediction in a setting where each method is tasked to classify disease relations based on the disease knowledge graph alone. This setting evaluates the flexibility of REMAP as REMAP can answer either graph-based or text-based disease queries. In particular, Table~\ref{tab:result} shows performance results attained on the human-annotated dataset with query disease pairs given only as disease concept nodes in the knowledge graph. We found TuckER significantly outperforms other knowledge graph embedding baselines in both accuracy and F1 score. Last, we find that REMAP-B is a top performer among REMAP variants, achieving an accuracy of 89.8 and an F1-score of 84.1.



\section{Discussion}\label{sec:discussion}
Next, we analyze results focusing on the selection of negative disease-disease samples and the impact negative sampling has on model performance. We also examine trade-offs between translation and bilinear methods (Section \ref{sec:dis1}). Finally, we give a case study illustrating REMAP predictions (Section \ref{sec:dis2}) and provide an ablation study into key REMAP components (Section \ref{sec:ablation}).

\subsection{Further analysis of results}\label{sec:dis1}
Our dataset's negative samples are more representative than the previous approaches \cite{lin2020highthroughput} of generating negative samples. These negative samples are selected from the EHR co-occurrence matrix, which has the co-occurrence number of disease-disease pairs below the threshold we set. In this paper, the negative samples constructed in this way do not require conceptual replacement (that is, they are not “generated” negative samples), so the negative samples are more in line with the actual situation. At the same time, threshold control is also used to reduce the false negative rate. However, the negative samples constructed in this way make it more difficult to distinguish between traditional machine learning and deep learning models, so a more fusion-capable model is needed for mining. 

Further, we find that TuckER obtained better performance than TransE in the experiments. We hypothesize that this finding is due to the inherent limitation of TransE in the presence of 1-to-N relationships. Specifically, suppose multiple target diseases exist for a particular source disease and a relation. In that case, the representation returned by TransE fails to capture the local graph neighborhoods of all target diseases simultaneously. That is because $\mathbf{t} \approx \mathbf{h}+\mathbf{r}$, meaning that the model does not have sufficient capacity to differentiate between embeddings $\mathbf{t}$ for different target diseases. In contrast, bilinear representation of TuckER can address this limitation of TransE, which explains its better performance. For this reason, our REMAP approach uses the TuckER component to facilitate joint learning.

\subsection{Case study}\label{sec:dis2}
We proceed with a case study examining the May Cause (MC) relationship between hypobetalipoproteinemia and fatty liver disease to illustrate how REMAP uses both graph and text modalities to classify the MC type disease-disease relationship. Several studies~\cite{schonfeld2003fatty,rodrigues2016non} found the hypobetalipoproteinemia can cause fatty liver. Figure~\ref{Fig. 3} illustrates the prediction of the MC relationship between hypobetalipoproteinemia and fatty liver made by the REMAP-B joint learning model. The graph can provide graph structure information to REMAP-B. For example, hypobetalipoproteinemia has 8 outgoing edges of MC type, which are the most among all edge types, and fatty liver disease also has 4 outgoing edges of the May Be Caused (MBC) type. And the graph encoder can also capture information from any possible meta-paths linking these two nodes in a shape of Hypobetalipoproteinemia $\underrightarrow{MC}$ $X$ $\underrightarrow{MC}$ Fatty liver, where $X$ is also a node in the knowledge graph.
The language encoder can capture language information from the free text, semantic types, and spans of disease from special tokens we added to the vocabulary of the pretrained language model.
In the case of joint learning, the text-based model can extract part of the disease representation from the knowledge graph to update its internal representations and thus improve text-based classification of relations.

\subsection{Ablation study} \label{sec:ablation}

\begin{table}[tbp]
 \caption{\label{tab:ablation} \textbf{Results of the ablation study.} DDx: differential diagnosis, MC: may cause, MBC: may be caused by. The ``micro'' columns denote micro average accuracy or F1-score for DDx, MC, and MBC relation types.}
 \centering
  \setlength{\tabcolsep}{1.5mm}{
 \begin{tabular}{ll|cccc|cccc}
  \toprule
  \multirow{2}{*}{Modality} &\multirow{2}{*}{Model} & \multicolumn{4}{c}{Accuracy} & \multicolumn{4}{c}{F1-score} \\
   && micro & DDx & MC & MBC & micro & DDx & MC & MBC  \\
  \midrule
  \multirow{6}{*}{Text} 
  &REMAP-B & \textbf{88.6} & \textbf{84.4} & \textbf{89.2} & 92.4 &\textbf{81.8} & \textbf{81.9} & 80.3 & 83.3\\
  & w/o joint learning (linear) & 87.3 & 84.1 & 86.3 & 91.3 & 78.9 & 81.2 & 73.4 & 80.9 \\
  & w/o joint learning (TransE) & 86.1	&79.1	&89.1	&90.2 & 80.5 & 79.3 & \textbf{82.2} & 80.8\\
  & w/o joint learning (TuckER) & 87.9 & 83.6 & 87.0 & \textbf{93.2} & 80.0 & 80.1 & 75.7 & 85.1 \\
  & w/o EHR embedding & 87.6 & 83.2 & 88.2 & 91.4 & 79.6 & 79.8 & 78.2 & 81.1\\
  & w/o unaligned triplets & 88.2 & 83.0 & 88.5 & \textbf{93.2} & 81.0 & 80.0 & 78.8 & \textbf{85.2}\\
  \midrule
  \multirow{6}{*}{Graph} 
  &REMAP-B & \textbf{89.8} & \textbf{87.3} & \textbf{89.9} & \textbf{92.2} & \textbf{84.1} & \textbf{85.8} & \textbf{82.4} & \textbf{82.7}\\
  & w/o joint learning (linear) & 87.4 & 82.4 & 89.3 & 90.2 & 80.5 & 80.5 & 82.1 & 78.4\\
  & w/o joint learning (TransE) & 85.8 & 81.6 & 85.3 & 90.4 & 77.2 & 78.9 & 73.1 & 78.9 \\
  & w/o joint learning (TuckER) & 88.9 & 85.8 & 89.4 & 91.6 & 82.5 & 83.9 & 81.0 & 81.4 \\
  & w/o EHR embedding & 87.6 & 84.4 & 87.2 & 91.4 & 80.3 & 82.0 & 76.8 & 81.4\\
  & w/o unaligned triplets & 87.3 & 84.8 & 87.8 & 89.4 & 79.5 & 82.4 & 77.7 & 76.2\\
  \bottomrule
 \end{tabular}}
\end{table}

We conduct an ablation study to provide evidence for the effectiveness of four key components in REMAP-B: joint learning loss, scoring functions, EHR embedding, and unaligned triplets in the knowledge graph.
Table~\ref{tab:ablation} shows results that compare performance REMAP-B and performance after excluding each component.
We observe performance drops when excluding joint learning techniques in REMAP-B. This conclusion is consistent across almost all the metrics and edge types in both text and graph modality, except the edge type MBC (May Be Caused) in text modality, on which the accuracy increases 0.8 from 92.4 to 93.2 percent when excluding joint learning with TuckER.
We also find that EHR embedding plays a significant role in REMAP-B since the performance consistently decreases in all metrics, edge types, and both modalities. The largest drop in performance is observed on the F1-score in graph modality, which decreases by 3.8 percent from 84.1 to 80.3. This analysis demonstrates that REMAP-B can effectively employ the EHR data for disease relation extraction.

Further, accuracy and F1-score also decrease when we remove unaligned triplets in the knowledge graph in all settings except the edge types MBC (May Be Caused), on which the accuracy increases 0.8 percent and F1-score increases 1.9 percent. This result indicates the importance of leveraging unsupervised information from unaligned triplets.
In summary, all components we introduce in REMAP-B are vital for achieving outstanding performance on disease relation extraction. Among all these components, joint learning techniques are most important since performance drops considerably when we exclude them, and this result is consistent across all settings.

\section{Conclusion}\label{sec:conclusion}
We develop a multi-modal learning approach REMAP for disease relation extraction that fuses language modeling with knowledge graphs. Results on a dataset of clinical expert annotations show that REMAP considerably outperforms methods for learning on text or knowledge graphs alone. Further, REMAP can extract and classify disease relationships in the most challenging settings where text or graph information is absent. Finally, we provide a new data resource of extracted relationships between diseases that can serve as a benchmarking dataset for systematic evaluation and comparison of disease relation extraction algorithms.

\clearpage

\section*{Funding and acknowledgments}

We gratefully acknowledge the support of the Harvard Translational Data Science Center for a Learning Healthcare System (CELeHS). M.Z. is supported, in part, by NSF under nos. IIS-2030459 and IIS-2033384, US Air Force Contract No. FA8702-15-D-0001, Harvard Data Science Initiative, Amazon Research Award, Bayer Early Excellence in Science Award, AstraZeneca Research, and Roche Alliance with Distinguished Scientists Award. Any opinions, findings, conclusions or recommendations expressed in this material are those of the authors and do not necessarily reflect the views of the funders.

\section*{Author contributions}
Y.L and K.L. are co-first authors and have developed the method and carried out all analyses in this study. S.Y. contributed experimental data. T.C. and M.Z. conceived and designed the study. The project was guided by M.Z., including designing methodology and outlining experiments. Y.L., K.L., S.Y., T.C., and M.Z. wrote the manuscript. All authors discussed the results and reviewed the paper. 

\section*{Data and code availability}

Python implementation of REMAP is available on Github at 
\url{https://github.com/Lukeming-tsinghua/REMOD}.
The human annotated dataset used for evaluation of REMAP is available at \url{https://doi.org/10.6084/m9.figshare.17776865}.

\section*{Conflicts of interest}

None declared. 

\clearpage

\begin{figure}
\centering
\includegraphics[width=0.8\linewidth]{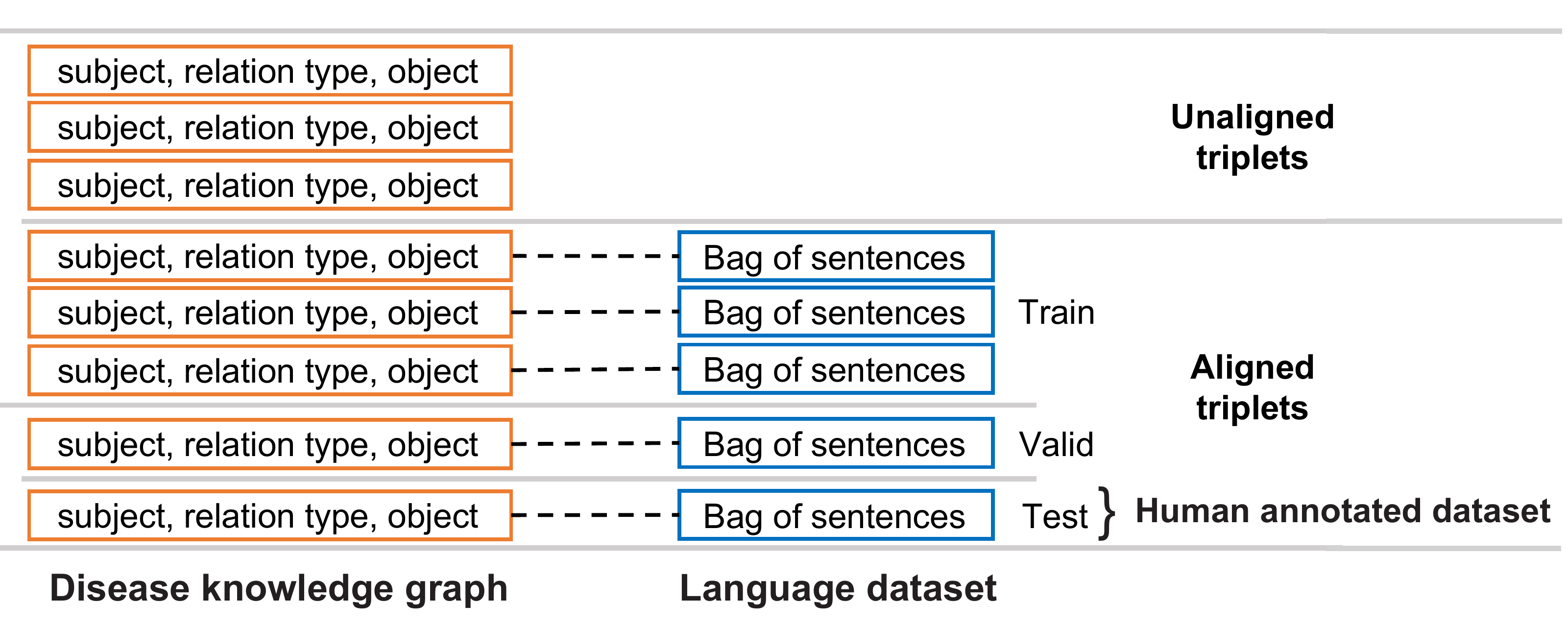} 
\caption{\textbf{Creating a dataset split for performance evaluation and benchmarking.} Triplets in the disease knowledge graph are divided into aligned triplets and unaligned triplets. {\em Aligned triplets} are triplets that have corresponding sentences in the language dataset. {\em Unaligned triplets} have no corresponding sentences in the language dataset.
}
\label{Fig2}
\end{figure}

\begin{figure}
\centering
\includegraphics[width=0.9\linewidth]{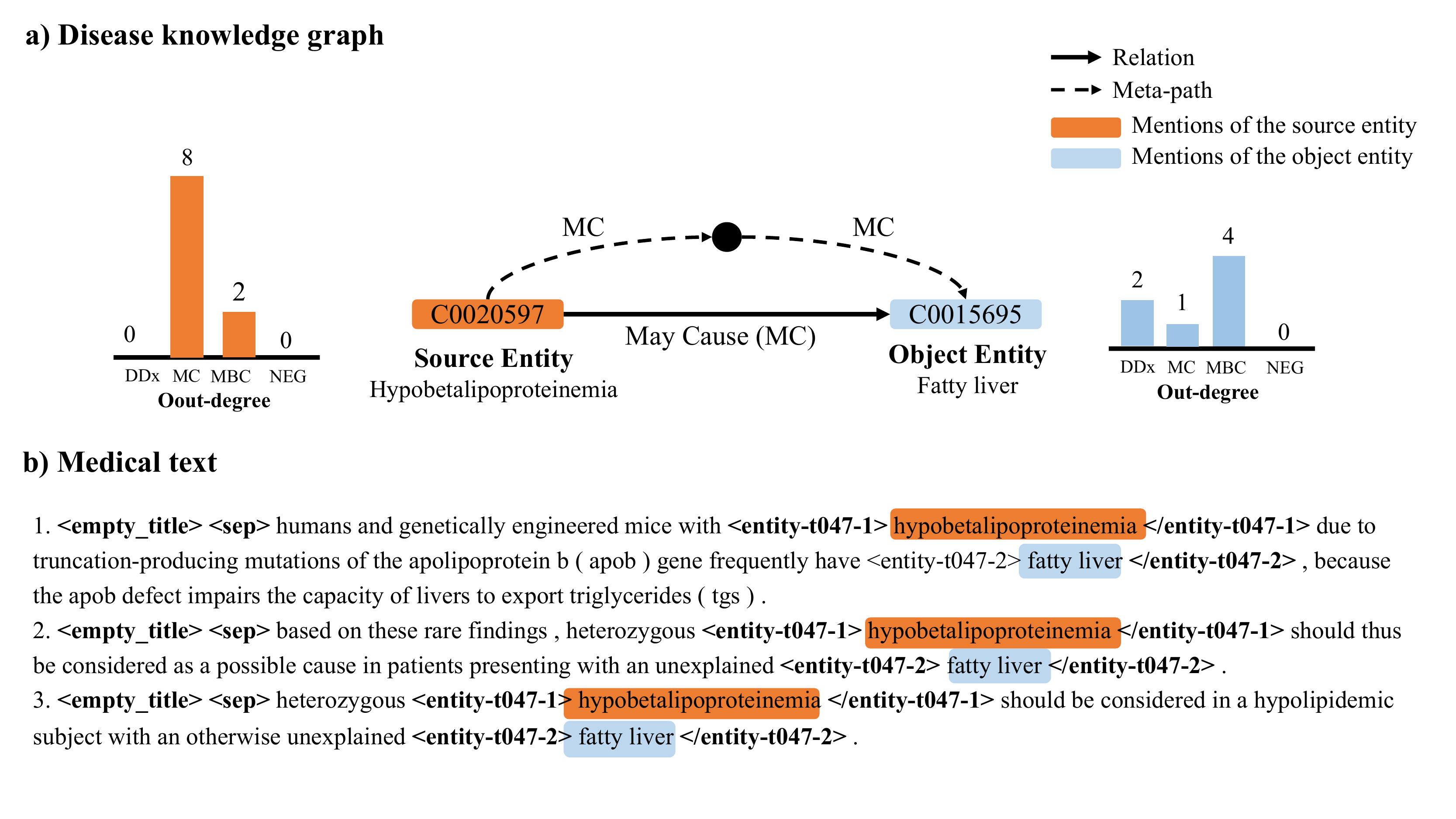} 
\caption{\textbf{Illustration of (Hypobetalipoproteinemia, May Cause, Fatty liver) triplet in REMAP.} This triplet represents the Hypobetalipoproteinemia has the symptom of fatty liver. The subject entity (Hypobetalipoproteinemia, C0020597) is shown in orange and the object entity (Fatty liver, C0015695) is shown in blue. \textbf{(a)}
The subject and object entities are identified with unique UMLS concept identifiers. The relation between them is the may cause (MC). There is a MC-MC meta-path between these entities and that information is leveraged by our graph encoder to predict the relationship between Hypobetalipoproteinemia and Fatty liver. The bar plots indicate distribution of relation types going out of the subject or object entities in the knowledge graph. \textbf{(b)} We show three sentences representing the triplet. We use \textit{<sep>} token to separate the sentences and the title of article from which we mine the sentence. If the article is from PubMed, we use special token \textit{<empty\_title>} as a placeholder. We add two special tokens before and after terms in each sentence to identify the positions of entities. For example, we add \textit{<entity-t047-1>} before \textit{Hypobetalipoproteinemia} to mark the beginning of subject entity; \textit{t047} serves as a identifier of the semantic type - \textit{Disease or Syndrome} - for Hypobetalipoproteinemia.}
\label{Fig. 3}
\end{figure}

\clearpage

\appendix
\setcounter{table}{0} \renewcommand{\thetable}{A\arabic{table}}

\section{Data preprocessing}\label{sec:data_preprocessing}

Data preprocessing and feature engineering follow the strategy utlined in Lin \etal \cite{lin2020highthroughput}. We outline the data preparation process in this section. This process transforms the raw datasets, including relation triplets, text corpus, and electronic health records embeddings into the AI-ready data for AI analyses. The code is publicly available on GitHub\footnote{\url{https://github.com/lychyzclc/High-throughput-relation-extraction-algorithm}}.

\subsection{Relation triplet collection}

Acquiring relation triplets is the first step in the preparation of training data. The majority of disease relation triplets are directly collected from the Diseases Database~\cite{pletscher2015diseases}. 
We also collect relation triplets by resolving semi-structured content on the MedScape.  Pages on it usually follow a very standard template, which allows one to easily locate sections about the target relations, where the entities are usually presented in lists and tables. Most commonly, the page title provides the head entity, the section title specifies the relation, lists, and tables in the section provide the tail entities. Therefore, we write simple web scraping scripts and apply maximum mapping to identify mentions of entities with UMLS CUIs, and assemble them into relation triplets.

\subsection{Document preparation}

The input to this step is a set of documents collected from online sources. These documents are used to pretrain a model \textit{en\_core\_web\_sm} in \textit{SpaCy} to split the document into sentences with additional structure information, including the title and subject headings.

\subsection{Entity linking}

The input to entity linking are sentences extracted from corpus. We use the nested forward maximum matching algorithm to annotate the mentions with medical terms collected from UMLS.

\subsection{Data cleaning}

We discard sentences that are shorter than 5 words. And we generate the static position embedding for baseline models, such as TextCNN and RNNs. We also make sure the head entity appears before the tail entity in sentences, and those sentences with entities in reverse order are moved to the opposite relation (i.e., from may\_cause to may\_be\_caused\_by).

\section{Further information on REMAP's performance}\label{sec:further-performance}

Table \ref{tab:result-text} extends Table \ref{tab:result} from the main text with additional information about REMAP's performance measured using precision, recall, and F1-score metrics. Shown is performance that REMAP achieves when the REMAP model, although trained on the multimodal graph-text dataset, is asked to make predictions at test time using only text information. 

Table \ref{tab:result-graph} extends Table \ref{tab:result} from the main text with additional information about REMAP's performance measured using precision, recall, and F1-score metrics. Shown is performance that REMAP achieves when the REMAP model, although trained on the multimodal graph-text dataset, is asked to make predictions at test time using only information from the disease knowledge graph. 

\begin{table}[t]
 \caption{\label{tab:result-text} \textbf{Performance of multimodal methods for identifying candidate disease-disease relations \underline{from text data.}} Shown is average performance across multiple independent runs calculated on the human annotated set. Higher values indicate better performance.}
 \centering
 \footnotesize
 \setlength{\tabcolsep}{1.7mm}{
 \begin{tabular}{lcccc|cccc|cccc}
  \toprule
  \multirow{2}{*}{Model} & \multicolumn{4}{c}{Precision} & \multicolumn{4}{c}{Recall}  & \multicolumn{4}{c}{F1-score} \\
   & minor & DDx & MC & MBC & minor & DDx & MC & MBC & minor & DDx & MC & MBC \\
   \midrule
    Random Forest &	69.2 & 71.8	& 67.6 & 58.3 & 26.0 & 42.2 & 15.7 & 58.3 & 37.8 & 53.1 & 25.5 & 19.2\\
    TextCNN	&67.8	&76.2	&56.7	&78.2	&55.3	&60.7	&63.5	&35.2	&60.9	&67.5	&59.9	&48.6\\
    BiGRU	&69.1	&68.1	&75.3	&65.7	&56.3	&67.8	&42.1	&54.9	&62.0	&67.9	&54.0	&59.8\\
    BiGRU+attention	&70.6	&74.4	&67.9	&67.7	&59.6	&62.1	&59.7	&54.9	&64.6	&67.7	&63.5	&60.6\\
    PubmedBERT	&71.2	& 64.0	&71.2	& 78.4	&87.4	&91.2	&85.5	&85.3	&78.5	&75.2	&74.6	&81.7 \\
  \midrule
  SciBERT (linear) & 86.6 & 81.0 & 96.9 & 88.3 & 72.5 & 81.4 & 59.1 & 74.6 & 78.9 & 81.2 & 73.4 & 80.9 \\
 SciBERT (TransE)  & 74.9 & 68.2 & 86.2 & 77.4 & 87.0 & 94.8 & 78.6 &	84.4 & 80.5 & 79.3 & 82.2 & 80.8 \\
 SciBERT (TuckER) & 87.3 & \textbf{81.7} & 93.5 & 91.5 & 73.9 & 78.6 & 63.5 & \textbf{79.5} & 80.0 & 80.1 & 75.7 & \textbf{85.1} \\
  \midrule
 REMAP & 85.8 & 79.9 & 94.8 & 88.0 & 76.6 & 81.4 & \textbf{69.2} & 77.9 & 80.9 & 80.7 & 80.0 & 82.6 \\
 REMAP-M & \textbf{87.4} & 79.9 & \textbf{97.3} & \textbf{93.0} & 76.4 & 83.3 & 67.3 & 76.2 & 81.5 & 81.6 & 79.6 & 83.8 \\
 REMAP-B & 86.2 & 79.7 & 95.7 & 89.6 & \textbf{77.8} & \textbf{84.3} & \textbf{69.2} & 77.9 &\textbf{81.8} & \textbf{81.9} & \textbf{80.3} & 83.3\\
  \bottomrule
 \end{tabular}}
\end{table}

\begin{table}[t]
 \caption{\label{tab:result-graph}\textbf{Performance of multimodal methods for identifying candidate disease-disease relations \underline{from graph-structured data.}} Shown is average performance across multiple independent runs calculated on the human annotated set. Higher values indicate better performance.}
 \centering
 \footnotesize
 \begin{tabular}{lcccc|cccc|cccc}
  \toprule
  \multirow{2}{*}{Model} & \multicolumn{4}{c}{Precision} & \multicolumn{4}{c}{Recall}  & \multicolumn{4}{c}{F1-score} \\
   & minor & DDx & MC & MBC & minor & DDx & MC & MBC & minor & DDx & MC & MBC \\
  \midrule
  TransE\_l2 & 61.3 & 63.0 & 57.3 & 63.0 & 65.2 & 73.8 & 56.6 & 61.5 & 63.2 & 68.0 & 57.0 & 62.2 \\
  DistMult & 53.6 & 77.8 & 40.7 & 43.1 & 58.9 & 65.2 & 46.5 & 63.9 & 56.1 & 71.0 & 43.4 & 51.5 \\
  ComplEx & 71.7 & 72.2 & 75.0 & 68.4 & 59.5 & 66.7 & 45.3 & 65.6 & 65.0 & 69.3 & 56.5 & 66.9 \\
  RGCN & 56.2 & 74.9 & 44.2 & 48.9 & 69.7 & 75.2 & 59.7 & 73.0 & 62.2 & 75.1 & 50.8 & 58.6 \\
  TuckER & 70.1 & 69.8 & 74.3 & 66.2 & 77.8 & 83.8 & 69.2 & 78.7 & 73.7 & 76.2 & 71.7 & 71.9 \\
  \midrule
  HAN (linear) & 82.0 & 75.2 & 92.9 & 84.8 & 79.0 & 86.7 & \textbf{73.6} & 73.0 & 80.5 & 80.5 & \textbf{82.1} & 78.4 \\
  HAN (TransE) & 81.3 & 76.1 & 88.4 & 84.9 & 73.5 & 81.9 & 62.3 & 73.8 & 77.2 & 78.9 & 73.1 & 78.9 \\
  HAN (TuckER) & 85.7 & 80.1 & 94.2 & 88.5 & 79.4 & 88.1 & 71.1 & 75.4 & 82.5 & 83.9 & 81.0 & 81.4 \\
  \midrule
  REMAP & \textbf{87.0} & \textbf{81.7} & 93.5 & \textbf{90.6} & 80.2 & 87.1 & 72.3 & \textbf{78.7} & 83.5 & 84.3 & 81.6 & \textbf{84.2} \\
  REMAP-M & 85.6 & 79.8 & \textbf{93.9} & 89.7 & 81.1 & \textbf{92.4} & 67.9 & \textbf{78.7} & 83.3 & 85.7 & 78.8 & 83.8 \\
  REMAP-B & 86.6 & 81.3 & 93.6 & 90.3 & \textbf{81.7} & 91.0 & \textbf{73.6} & 76.2 & \textbf{84.1} & \textbf{85.8} & \textbf{82.4} & 82.7\\
  \bottomrule
 \end{tabular}
\end{table}

\section{Hyper parameters}

\begin{table}[htbp]
 \caption{\label{tab:param} \textbf{Selection of hyper-parameters in REMAP}. We use grid search to select hyper-parameters on the validation set.}
 \centering
 \begin{tabular}{lllc}
  \toprule
  REMAP's component & Model parameter & Notation & Value \\
  \midrule
    \multirow{6}{*}{Neural architecture}
    &Padding length of sentences & $d_l$ & 256\\
    &Hidden size of SciBERT output & $d_{hs}$ & 768 \\
    &Hidden size of HAN output & $d_{ha}$  &100 \\
    &Hidden size of initial node embedding & $d_{hi}$ & 1,000\\
    &Hidden size of node embedding & $d_h$ & 100\\
    &Hidden size of relation embedding & $d_r$ & 100\\
  \midrule
  \multirow{9}{*}{Multimodal training}
  &Max sentence sample number & $l_{m(\max)}$ & 12 \\
  &Training batch size & $b_{\textrm{train}}$ & 4\\
  &Test batch size & $b_{\textrm{test}}$ & 16\\
  &Weight decay & $wd$ &$5\times 10^{-5}$ \\
  &Learning rate & $lr$ & $1\times 10^{-5}$\\
  &Gradient accumulate step & $\textrm{step}_g$ & 4\\
  &Optimizer & & Adam\\
  &Scheduler & & Linear \\
  &Warmup rate & $r_{\textrm{warmup}}$ & 0.1\\
  \midrule
  \multirow{9}{*}{Language model}
  &Max sentence sample number & $l_{m(\max)}$ & 12 \\
  &Training batch size & $b_{\textrm{train}}$ & 4\\
  &Test batch size & $b_{\textrm{test}}$ & 16\\
  &Weight decay & $wd$ &$5\times 10^{-5}$ \\
  &Learning rate & $lr$ & $1\times 10^{-5}$\\
  &Gradient accumulate step & $\textrm{step}_g$ & 4\\
  &Optimizer & & Adam\\
  &Scheduler & & Linear\\
  &Warmup rate & $r_{\textrm{warmup}}$ & 0.1\\
  \midrule
  \multirow{7}{*}{Graph model}
  &Training batch size & $b_{\textrm{train}}$ & 512\\
  &Test batch size & $b_{\textrm{test}}$ & 512\\
  &Weight decay & $wd$ & $1\times 10^{-8}$\\
  &Learning rate & $lr$ & $1\times 10^{-3}$\\
  &Optimizer & & Adam \\
  &Scheduler & & StepLR\\
  & StepLR scheduler & $\gamma$ & 0.9\\
  \bottomrule
 \end{tabular}
\end{table}

\clearpage

\bibliographystyle{elsarticle-num} 
\bibliography{ref.bib}





\end{document}